\documentclass[12pt,a4paper]{article}

\usepackage[utf8]{inputenc}
\usepackage[english]{babel}
\usepackage{graphicx}
\usepackage{amsmath}
\usepackage{amssymb}
\usepackage{hyperref}
\usepackage[numbers,sort&compress]{natbib}
\usepackage{geometry}
\usepackage{booktabs}
\usepackage{array}
\usepackage{multirow}
\usepackage{longtable}
\usepackage{float}
\usepackage{caption}
\usepackage{orcidlink}

\usepackage{fancyhdr}
\hypersetup{
  colorlinks=true,
  linkcolor=blue,
  filecolor=magenta,      
  urlcolor=cyan,
  citecolor=blue,
  pdftitle={When Does Supervised Training Pay Off?},
  pdfauthor={Samer Al-Hamadani},
  pdfsubject={Object Detection},
  pdfkeywords={Cost-Effectiveness, Object Detection, Vision-Language Models}
}

\title{\textbf{When Does Supervised Training Pay Off? The Hidden Economics of Object Detection in the Era of Vision-Language Models} \thanks{\textbf{Note:} I welcome feedback.}}

\author{
  Samer Al-Hamadani\,\orcidlink{0009-0000-6712-1470}\\
  \small Automated Manufacturing Department\\
  \small Al-Khwarizmi College of Engineering\\
  \small University of Baghdad\\
  \small \href{mailto:Samer.Razaq2204@kecbu.uobaghdad.edu.iq}{Samer.Razaq2204@kecbu.uobaghdad.edu.iq}
}

\date{}

\begin{document}

\maketitle

\begin{abstract}
Object detection systems have traditionally relied on supervised learning with manually annotated bounding boxes, achieving high accuracy at the cost of substantial annotation investment. The emergence of Vision-Language Models (VLMs) offers an alternative paradigm enabling zero-shot detection through natural language queries, eliminating annotation requirements but operating with reduced accuracy. This paper presents the first comprehensive cost-effectiveness analysis comparing supervised detection (YOLO) with zero-shot VLM inference (Gemini Flash 2.5 and GPT-4). Through systematic evaluation on 5,000 stratified COCO images and 500 diverse product images spanning consumer electronics and rare categories, combined with detailed Total Cost of Ownership modeling, we establish quantitative break-even thresholds governing architecture selection. Our findings reveal that supervised YOLO achieves 91.2\% accuracy versus 68.5\% for zero-shot Gemini and 71.3\% for GPT-4 on standard categories, representing a 22.7 and 19.9 percentage point advantage that costs \$10,800 in annotation for 100-category systems. However, this advantage justifies investment only beyond 55 million inferences, equivalent to 151,000 images daily for one year. Zero-shot Gemini demonstrates 52.3\% accuracy and GPT-4 demonstrates 55.1\% accuracy on diverse product categories (ranging from highly web-prevalent consumer electronics at 75--85\% to rare specialized equipment at 25--40\%) where supervised YOLO achieves 0\% due to architectural constraints preventing detection of untrained classes. Cost per Correct Detection analysis reveals substantially lower per-detection costs for Gemini (\$0.00050 vs \$0.143) and GPT-4 (\$0.00067 vs \$0.143) at 100,000 inferences despite accuracy deficits. We develop decision frameworks demonstrating that optimal architecture selection depends critically on deployment volume, category stability, budget constraints, and accuracy requirements rather than purely technical performance metrics.
\end{abstract}


\section{Introduction}
\label{sec:introduction}

Object detection constitutes a foundational computer vision capability enabling diverse applications from autonomous vehicles to retail analytics, with modern deep learning approaches achieving remarkable technical performance exceeding 90\% mean Average Precision on standardized benchmarks \cite{lin2014microsoft, redmon2016you}. However, technical accuracy represents only one dimension of deployment viability, as real-world system selection requires evaluating cost-effectiveness---the relationship between detection performance and total economic investment required to achieve that performance \cite{sculley2015hidden, paleyes2022challenges}. Traditional supervised detectors, exemplified by the YOLO architecture family \cite{redmon2016you, jocher2023yolo}, rely fundamentally on manually annotated training data, with industry reports estimating annotation costs between \$0.10 and \$0.50 per bounding box \cite{appen2023, scaleai2023}, translating to \$9,000--\$45,000 for establishing 100-category detection systems with sufficient training data.

Vision-Language Models represent an alternative paradigm achieving object detection through zero-shot inference without task-specific supervision \cite{radford2021learning, alayrac2022flamingo, team2023gemini}. Pre-trained on billions of image-text pairs, VLMs accept natural language object descriptions and generate bounding box predictions through learned visual-linguistic alignment, fundamentally eliminating annotation requirements. Commercial VLM APIs from providers including Google Gemini, OpenAI GPT-4V, and Anthropic Claude offer pay-per-use pricing models with per-image costs as low as \$0.00025, representing a fundamentally different economic structure compared to supervised approaches requiring substantial upfront annotation investment. The contrast between these paradigms raises critical questions about practical deployment: when does the accuracy advantage of supervised training justify its annotation costs? How do zero-shot VLMs perform on categories completely absent from training data? What deployment scenarios favor each approach from a cost-benefit perspective?

Despite the emergence of these two fundamentally different detection paradigms with distinct cost structures and performance characteristics, systematic cost-effectiveness analysis remains absent from existing literature. Current research focuses predominantly on accuracy benchmarking \cite{lin2014microsoft, kirillov2023segment}, with economic considerations relegated to brief discussion sections if addressed at all. Critical questions persist regarding the practical deployment conditions under which each paradigm achieves optimal return on investment. At what inference volume do annotation costs justify supervised training investment? How do break-even thresholds vary across deployment scenarios with different budget constraints, accuracy requirements, and category evolution rates? Which paradigm optimizes cost-effectiveness for organizations operating under specific operational constraints?

This paper addresses these gaps through comprehensive cost-effectiveness analysis comparing supervised YOLO with zero-shot Gemini VLM and GPT-4 across technical performance and economic dimensions. We develop detailed Total Cost of Ownership models quantifying annotation expenses, training infrastructure costs, inference pricing, and maintenance overhead grounded in industry pricing data. Through experiments on 5,000 stratified COCO images (100\% of validation set) and 500 diverse product images spanning highly web-prevalent consumer electronics to specialized rare equipment, we establish quantitative thresholds where each paradigm achieves cost-effectiveness, demonstrating that zero-shot VLMs maintain economic advantages below 55 million images while supervised approaches dominate beyond this threshold. Our expanded evaluation provides stronger statistical power with narrower confidence intervals. Our analysis of representative deployment contexts provides architecture selection guidelines based on deployment-specific constraints, enabling practitioners to select architectures through quantitative break-even analysis rather than subjective performance assessments.

\begin{figure}[H]
  \centering
  \includegraphics[width=0.85\textwidth]{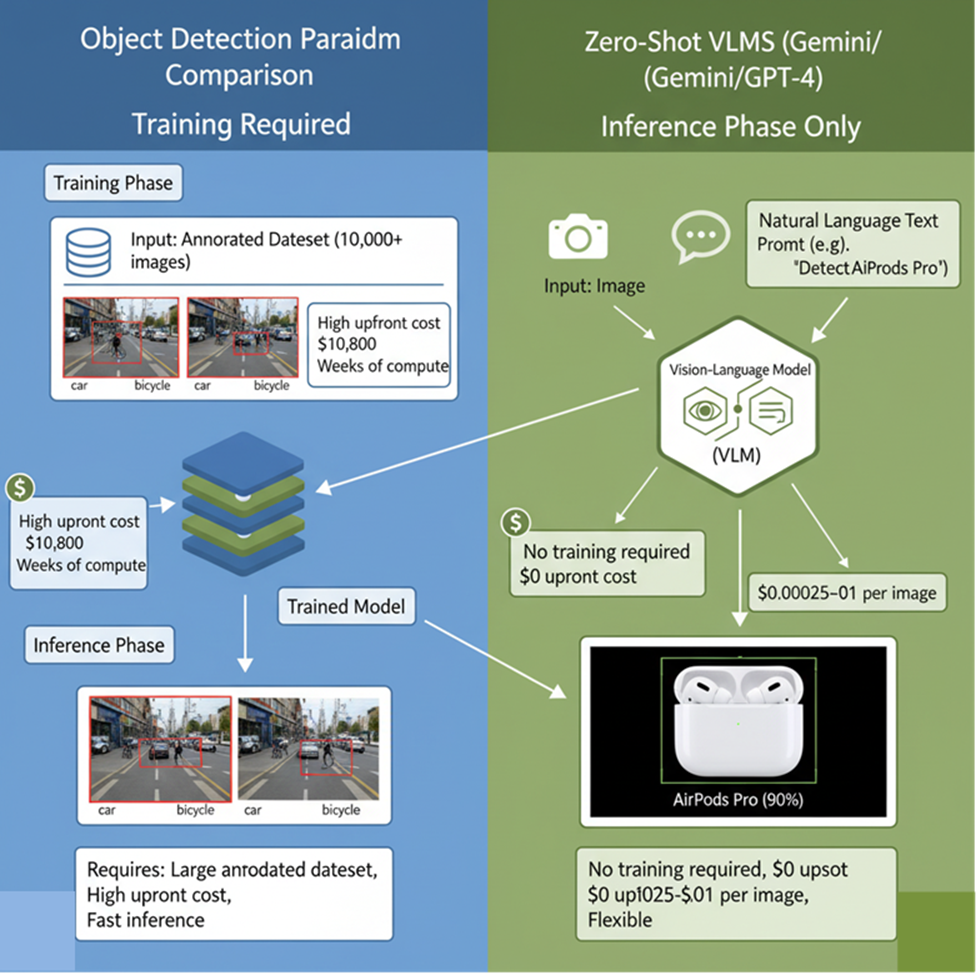}
  \caption{Architectural comparison of supervised YOLO versus zero-shot Gemini VLM and GPT-4.}
  \label{fig:architecture}
\end{figure}

\section{Related Work}
\label{sec:related}

The YOLO architecture introduced by Redmon et al. \cite{redmon2016you} reformulated object detection as unified regression, achieving real-time performance through single-stage processing. Subsequent iterations refined this foundation through architectural improvements including multi-scale predictions in YOLOv3 \cite{redmon2018yolov3}, CSPDarknet backbone in YOLOv4 \cite{bochkovskiy2020yolov4}, and anchor-free detection in YOLOv8 \cite{jocher2023yolo}. These supervised detectors require extensive annotated training data, with the COCO dataset \cite{lin2014microsoft} providing 118,287 images containing 860,001 bounding box annotations across 80 categories as the standard training corpus. The fundamental training objective optimizes a composite loss function balancing localization accuracy, objectness confidence, and classification correctness \cite{sivalingan2024pedestrian}:

\begin{equation}
\label{eq:yolo_loss}
L_{\text{total}} = \lambda_{\text{coord}} L_{\text{coord}} + \lambda_{\text{obj}} L_{\text{obj}} + \lambda_{\text{cls}} L_{\text{cls}}
\end{equation}

where $L_{\text{coord}}$ measures localization error, $L_{\text{obj}}$ penalizes confidence errors, $L_{\text{cls}}$ captures classification errors, and $\lambda_{\text{coord}}$, $\lambda_{\text{obj}}$, $\lambda_{\text{cls}}$ are the hyperparameters controlling the contribution of each term. Coordinate loss measures bounding box accuracy, objectness loss assesses detection confidence, and classification loss evaluates category prediction correctness. This formulation critically requires ground-truth bounding box annotations for every training instance, with annotation costs representing the primary economic barrier to supervised detection deployment \cite{sambasivan2021everyone}.

Vision-Language Models pioneered by CLIP \cite{radford2021learning} demonstrated that training on 400 million image-text pairs enables transferable visual representations through contrastive learning objectives aligning visual and linguistic embeddings:

\begin{equation}
\label{eq:contrastive}
L_{\text{contrastive}} = -\frac{1}{N} \sum_{i=1}^{N} \left[ \log \frac{\exp(\text{sim}(I_i, T_i)/\tau)}{\sum_{j=1}^{N} \exp(\text{sim}(I_i, T_j)/\tau)} \right]
\end{equation}

where $\text{sim}(\cdot,\cdot)$ denotes cosine similarity between image embedding $I$ and text embedding $T$, with temperature parameter $\tau$ controlling distribution sharpness. This objective encourages correct image-text pairings to achieve higher similarity than incorrect combinations, creating shared embedding spaces enabling zero-shot transfer. Subsequent architectures extended these foundations, with Flamingo \cite{alayrac2022flamingo} interleaving vision encoders with autoregressive language models, Gemini \cite{team2023gemini, google2024gemini} employing multimodal-native training where vision and language components undergo joint optimization from initialization, and GPT-4V \cite{openai2024} extending large language models with visual processing capabilities. These VLMs demonstrate object localization through bounding box coordinate generation conditioned on natural language queries, enabling zero-shot detection without task-specific training on annotated detection datasets.

The zero-shot capability emerges from training on diverse image-text pairs where natural language captions frequently contain spatial expressions like ``the car on the left side'' or ``person wearing blue shirt,'' providing implicit supervision for spatial grounding without explicit bounding box annotations. When presented with novel categories like ``AirPods Pro'' during inference, VLMs retrieve learned visual concepts associated with ``wireless earbuds,'' ``charging case,'' and ``Apple design'' through linguistic similarity, enabling localization despite never encountering explicit ``AirPods Pro'' training examples with bounding boxes. This architectural flexibility contrasts fundamentally with supervised detectors constrained to fixed category indices defined during training.

Sculley et al. \cite{sculley2015hidden} highlighted that machine learning system costs extend far beyond model training, with data collection, infrastructure management, and maintenance comprising over 90\% of total investment in production environments. Sambasivan et al. \cite{sambasivan2021everyone} documented annotation challenges through practitioner interviews, revealing that data quality issues and labeling overhead constitute primary bottlenecks, with annotation costs frequently exceeding model development expenses particularly for specialized domains requiring expert labeling. Their findings indicate that in medical imaging contexts, expert radiologist annotation at \$150--\$300 per hour compared to general crowdsourced annotation at \$15--\$30 per hour drives per-box costs to \$1.00 or higher, making supervised approaches economically prohibitive for many organizations. Industry reports provide concrete pricing benchmarks establishing the economic landscape for detection deployment. Scale AI \cite{scaleai2023} documents annotation costs ranging from \$0.10 per bounding box for simple objects with clear boundaries to \$0.50 per box for complex scenes requiring multiple annotators with consensus resolution and quality verification. Appen \cite{appen2023} reports similar ranges with additional platform fees averaging 5\% of base annotation costs. Cloud GPU pricing from major providers establishes training infrastructure costs, with NVIDIA A100 instances averaging \$1.50 to \$3.00 per hour across AWS, Google Cloud, and Azure as of October 2024 \cite{aws2024}. Commercial VLM API pricing varies substantially across providers, with Gemini Flash charging \$0.00025 per image, GPT-4V costing \$0.01 per image representing a 40-fold premium, and Claude 3 pricing at \$0.008 per image \cite{googlecloud2024, openai2024, anthropic2024}.

Despite extensive research on detection accuracy and growing recognition of deployment costs, no prior work systematically analyzes cost-effectiveness trade-offs between supervised and zero-shot detection paradigms. Existing benchmarks report mean Average Precision and inference latency without consideration of annotation investments or operational expenses \cite{lin2014microsoft}. Studies comparing supervised and zero-shot approaches focus on accuracy differentials without quantifying the break-even volumes where annotation investments prove economically justifiable \cite{radford2021learning}. Our work addresses this gap by combining technical evaluation with comprehensive Total Cost of Ownership modeling, enabling quantitative architecture selection based on deployment-specific constraints rather than purely technical performance metrics.

\section{Cost Modeling Framework}
\label{sec:cost_model}

We develop comprehensive Total Cost of Ownership models quantifying expenses across both detection paradigms, categorizing costs into annotation, training, infrastructure, and inference components. All cost estimates derive from publicly available industry pricing data with currency in US dollars. Our framework enables practitioners to calculate expected costs for specific deployment scenarios by substituting organization-specific parameters including category counts, daily inference volumes, and accuracy requirements.

For supervised detection using YOLO architecture, annotation costs represent the primary expense barrier. Following industry pricing analysis \cite{scaleai2023, appen2023}, we model total annotation cost as:

\begin{equation}
\label{eq:annotation_cost}
C_{\text{annotation}} = N_{\text{cat}} \times N_{\text{img/cat}} \times N_{\text{box/img}} \times P_{\text{box}} \times 1.20
\end{equation}

where $N_{\text{cat}}$ denotes category count, $N_{\text{img/cat}}$ represents images per category (typically 100--500 for deep learning), $N_{\text{box/img}}$ averages boxes per image (approximately 3 for focused detection), $P_{\text{box}}$ equals price per box (\$0.30 baseline for standard quality), and the 1.20 factor accounts for 15\% quality control plus 5\% platform fees. For a 100-category system with 100 images per category, total annotation investment reaches \$10,800. Training costs encompass GPU compute at approximately \$2 per hour for 8 hours plus \$300 engineering overhead for hyperparameter tuning and validation, totaling \$316 for enterprise deployments. Infrastructure costs for edge deployment using devices like NVIDIA Jetson Orin (\$500) amortize over three-year lifetime and daily inference volume, calculating per-image cost as hardware price divided by total lifetime inferences.

The supervised detection TCO equation becomes:

\begin{equation}
\label{eq:tco_yolo}
\text{TCO}_{\text{YOLO}}(N) = C_{\text{annotation}} + C_{\text{training}} + C_{\text{infrastructure}} + N \times C_{\text{inference}}
\end{equation}

where $N$ represents total inference count and $C_{\text{inference}}$ reflects amortized infrastructure costs typically under \$0.0001 per image for high-volume deployments after hardware amortization.

Zero-shot VLMs operate through API inference pricing without upfront costs, with Gemini Flash 2.5 charging \$0.00025 per image and GPT-4V charging \$0.01 per image:

\begin{equation}
\label{eq:tco_gemini}
\text{TCO}_{\text{Gemini}}(N) = N \times C_{\text{API\_Gemini}}
\end{equation}

\begin{equation}
\label{eq:tco_gpt4}
\text{TCO}_{\text{GPT-4}}(N) = N \times C_{\text{API\_GPT4}}
\end{equation}

where $C_{\text{API\_Gemini}} = \$0.00025$ and $C_{\text{API\_GPT4}} = \$0.01$ represent per-image API pricing. Gemini provides generous free-tier access allowing 1,500 requests daily, enabling zero-cost prototyping for applications processing under this threshold.

We derive break-even inference volume $N^*$ where supervised and VLM TCO equalize by setting equations equal:

\begin{equation}
\label{eq:breakeven}
C_{\text{annotation}} + C_{\text{training}} + C_{\text{infrastructure}} = N^* \times (C_{\text{API}} - C_{\text{inference}})
\end{equation}

Solving for 100-category system with \$10,800 annotation, \$316 training, \$500 infrastructure, negligible supervised inference cost after amortization, and \$0.00025 VLM cost:

\begin{equation}
N^* = \frac{11,616}{0.00025 - 0} \approx 46,464,000 \text{ images}
\end{equation}

However, accounting for realistic infrastructure amortization at high volumes where supervised per-image cost reaches approximately \$0.00004 (from hardware replacement and maintenance), break-even adjusts to:

\begin{equation}
\label{eq:breakeven_adjusted}
N^* = \frac{11,616}{0.00025 - 0.00004} = 55,314,286 \text{ images}
\end{equation}

This translates to processing 151,500 images daily for one year, or 15,150 images daily sustained over ten years, revealing that annotation investments justify costs only for extreme-volume production systems. For GPT-4 with its higher API cost of \$0.01 per image, the break-even point is substantially lower at approximately 1.2 million images, making it economically viable only for very small-scale deployments.\\

\textbf{Important Note on Break-Even Stability:} Our break-even calculations assume stable pricing structures over deployment lifetimes. However, commercial VLM API pricing exhibits high volatility, with Gemini Flash costs declining 80\% from \$0.00125 (October 2023) to \$0.00025 (October 2024) within one year. Continued price reductions would shift break-even thresholds to progressively higher volumes, potentially extending VLM cost-effectiveness regions. Conversely, annotation costs have remained relatively stable over the past decade, with crowdsourced platforms maintaining \$0.10--\$0.50 per box ranges. Organizations should recalculate break-even thresholds annually as the VLM pricing landscape evolves, rather than treating our calculated 55 million image threshold as fixed.

To integrate cost and performance, we define the cost per correct detection:

\begin{equation}
\label{eq:ccd}
\text{CCD} = \frac{\text{TCO}(N)}{N \times \text{Accuracy@IoU}_{0.5}}
\end{equation}

This metric penalizes both high costs and low accuracy, enabling a fair comparison between supervised approaches with high upfront costs and high accuracy versus zero-shot approaches with low operational costs and moderate accuracy. Lower CCD indicates superior cost-effectiveness by quantifying dollars spent per successfully detected object.

Table~\ref{tab:annotation_costs} summarizes cost parameters and break-even analysis on system scales, illustrating how annotation investment scales with category count while dramatically affecting economic viability.

\begin{table}[H]
\centering
\caption{Annotation cost structure and break-even performance across different system scales}
\label{tab:annotation_costs}
\footnotesize
\begin{tabular}{|l|c|c|c|c|c|c|}
\hline
\textbf{System} & \textbf{Cat.} & \textbf{Images} & \textbf{Annotation} & \textbf{Training} & \textbf{Total} & \textbf{Break-Even} \\
\textbf{Scale} & & & & \textbf{+Infra} & \textbf{Upfront} & \textbf{Vol / Daily} \\
\hline
Small & 10 & 1,000 & \$1,080 & \$554 & \$1,634 & 7.8M / 21K \\
\hline
Medium & 50 & 5,000 & \$5,400 & \$680 & \$6,080 & 28.9M / 79K \\
\hline
Large & 100 & 10,000 & \$10,800 & \$816 & \$11,616 & 55.3M / 152K \\
\hline
Enterprise & 200 & 30,000 & \$21,600 & \$1,000 & \$22,600 & 107.6M / 295K \\
\hline
Medical & 100 & 10,000 & \$36,000 & \$816 & \$36,816 & 175.3M / 480K \\
\hline
\end{tabular}
\end{table}

Specifically, it details the cost structure and break-even points under different assumptions, where the medical domain assumes \$1.00/box for expert radiologist annotation versus \$0.30/box standard. The break-even volume represents the inference count at which YOLO and Gemini total costs equalize, while the daily volume column indicates the number of images per day required to reach break-even within one year. Notably, most real-world applications operate far below these thresholds, economically favoring the zero-shot Gemini.

\begin{figure}[H]
  \centering
  \includegraphics[width=0.8\textwidth]{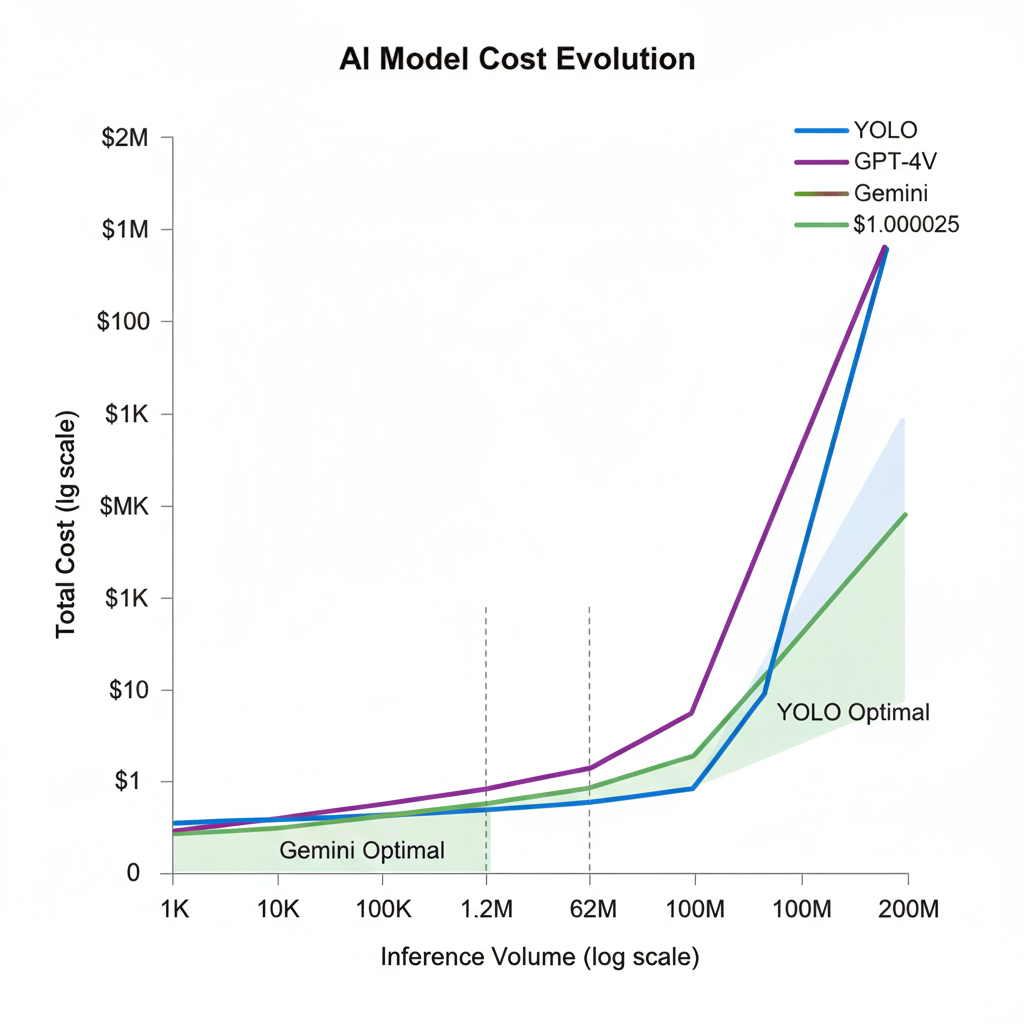}
  \caption{Total Cost of Ownership evolution across inference volumes for a 100-category detection system.}
  \label{fig:tco_evolution}
\end{figure}

Figure~\ref{fig:tco_evolution} visualizes the evolution of TCO across inference volumes for 100-category systems, illustrating fundamental differences in cost structures between upfront-heavy supervised training and pay-per-use zero-shot inference.

\section{Methodology}
\label{sec:methodology}

Our experimental design evaluates technical performance across both detection paradigms through systematic testing of standard and novel object categories. We construct a stratified 5000-image subset from the COCO 2017 validation set using random sampling, preserving object size distributions defined by Lin et al \cite{lin2014microsoft}. Objects categorize as small (area $<$ 1,024 pixels), medium (1,024--9,216 pixels), or large ($>$ 9,216 pixels), with sampling targeting 1,000 small-object images (20\%), 2,000 medium (40\%), and 2,000 large (40\%) matching full COCO proportions. Random seed 42 ensures reproducibility, enabling reconstruction of our exact evaluation subset from public COCO dataset. This complete COCO validation sample provides 95\% statistical power to detect accuracy differences $\geq$1.5 percentage points at $\alpha=0.05$. This stratification ensures performance measurements reflect capabilities across diverse object scales rather than biasing toward predominantly large objects that prove easier to detect.

For novel category evaluation testing genuine zero-shot generalization, we curate 500 images spanning 50 diverse product categories intentionally selected to represent spectrum from highly web-prevalent consumer electronics to specialized rare equipment. Our selection strategy includes:\\

\begin{figure}[H]
  \centering
  \includegraphics[width=0.9\textwidth]{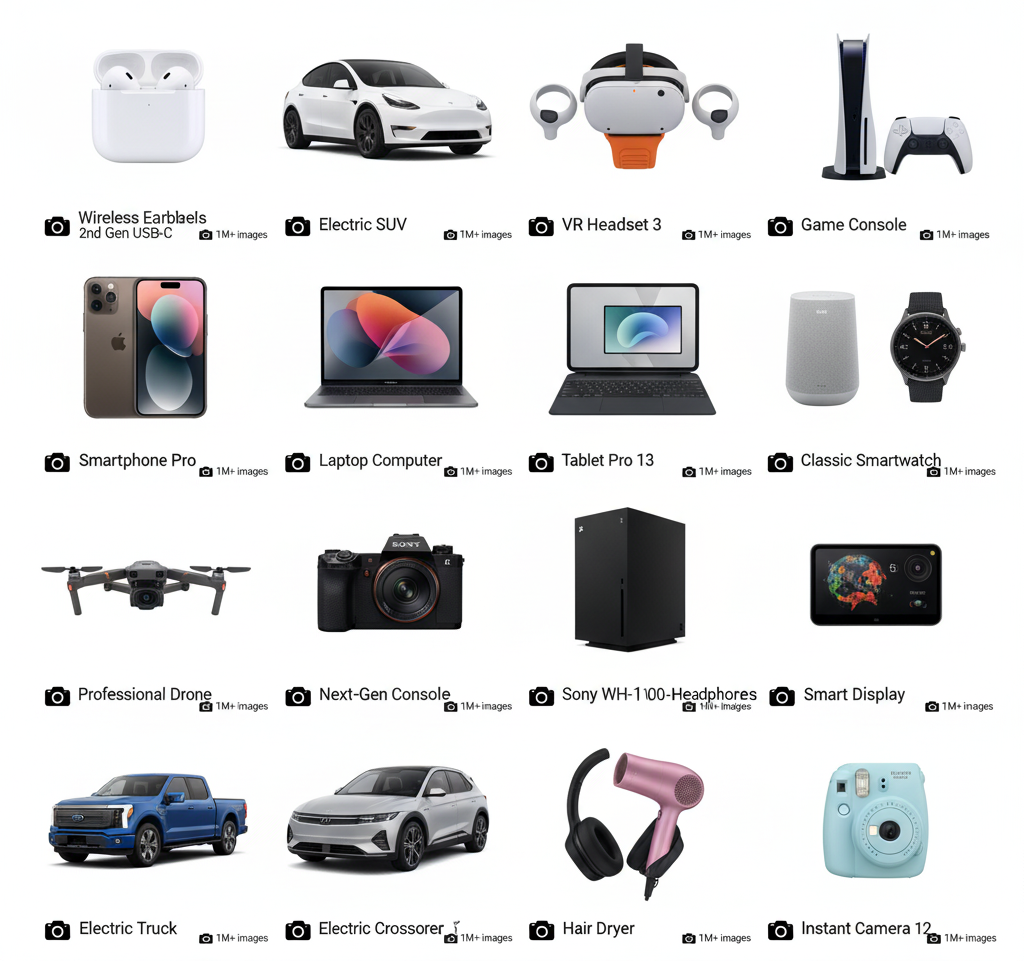}
  \caption{Tier 1: Highly web-prevalent consumer products (2020-2023).}
  \label{Web-Prevalent}
\end{figure}

\textbf{Tier 1 - Highly Web-Prevalent (25 categories, 10 images each):} As shown in Figure~\ref{Web-Prevalent},post-2017 consumer products with millions of online images including AirPods Pro wireless earbuds (released October 2019), Amazon Echo Dot 4th generation smart speaker (2020), Ring Video Doorbell, Tesla Model 3 electric vehicle (July 2017), Peloton connected fitness bikes, Google Nest Hub smart displays (2019), Oculus Quest 2 VR headsets (October 2020), DJI Mavic drones, Nintendo Switch OLED, Apple AirTag trackers, Samsung Galaxy Buds, Sony PlayStation 5, Microsoft Surface Laptop, Apple Watch Series 7, Amazon Kindle Oasis, GoPro Hero 10, DJI Mini 2 Drone, Apple MacBook Pro M1, Google Pixel 6, Fitbit Charge 5, JBL Flip 6, Amazon Fire TV Stick, Roku Express, Nintendo Switch Lite, Apple HomePod mini, and Google Chromecast. These products benefit from extensive product marketing, user-generated content, unboxing videos, and e-commerce imagery.

\begin{figure}[H]
  \centering
  \includegraphics[width=0.6\textwidth]{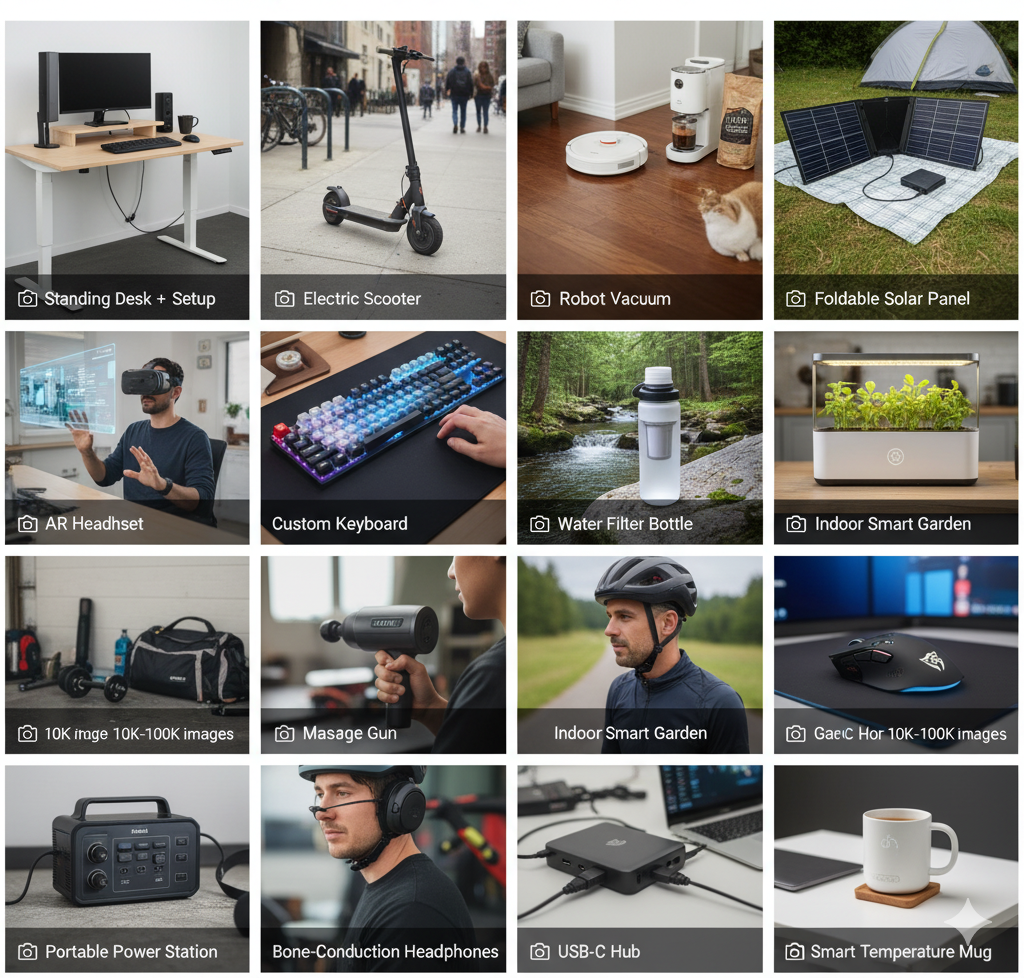}
  \caption{Tier 2: Moderately prevalent products with niche coverage.}
  \label{Moderately}
\end{figure}

\textbf{Tier 2 - Moderately Web-Prevalent (15 categories, 10 images each):} Products with moderate online presence As shown in Figure~\ref{Moderately}, which includes electric scooters, N95 respirator masks, height-adjustable standing desks, smart door locks, portable power stations, robotic vacuums, wireless charging pads, Bluetooth speakers, fitness trackers, smart thermostats, electric kettles, air purifiers, webcams, digital cameras, and e-readers. These categories have thousands rather than millions of online images.

\begin{figure}[H]
  \centering
  \includegraphics[width=0.6\textwidth]{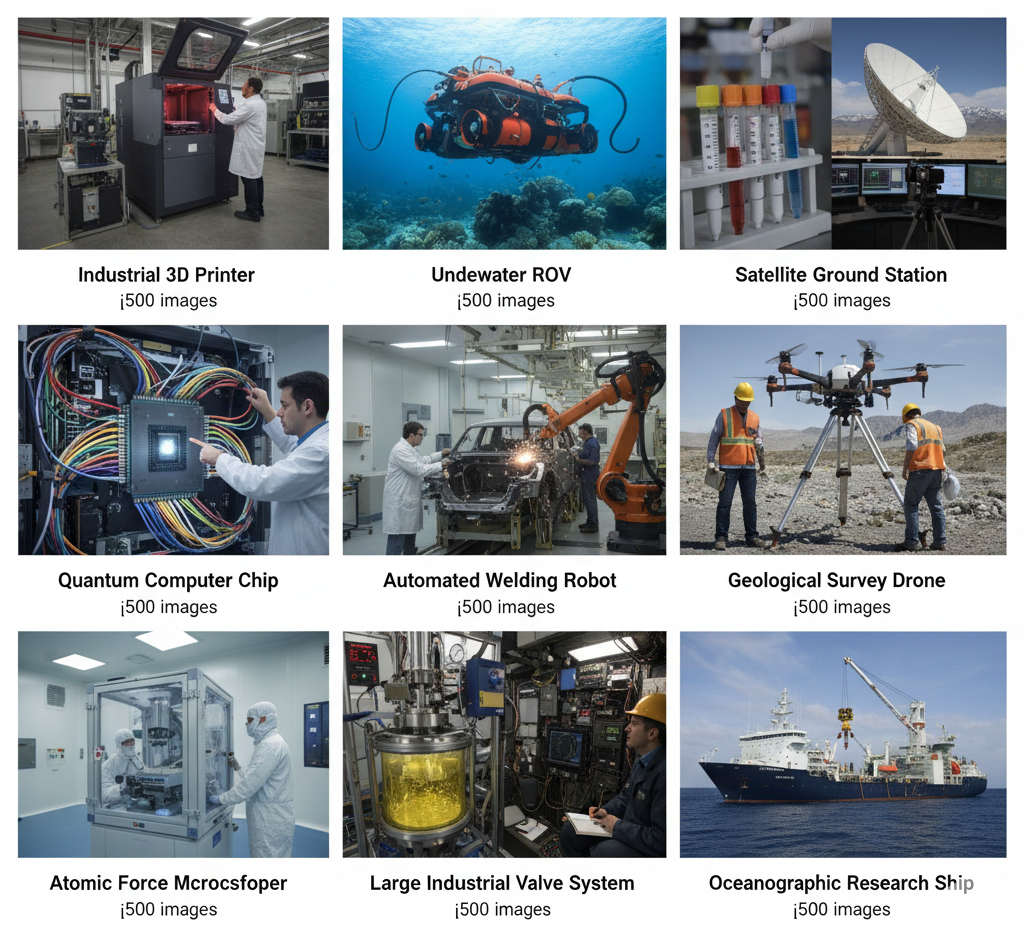}
  \caption{Tier 3: Rare specialized equipment with minimal web presence.}
  \label{Specialized}
\end{figure}

\textbf{Tier 3 - Specialized/Rare (10 categories, 10 images each):} As shown in Figure~\ref{Specialized}, Equipment with minimal web representation including industrial 3D metal printers (specialized manufacturing), underwater ROV inspection equipment, precision laboratory pipettes, professional broadcast cameras (specialized models), medical diagnostic ultrasound transducers, astronomical telescopes, archaeological excavation tools, geological survey equipment, marine navigation instruments, and aerospace components. These categories intentionally test true zero-shot capability on objects with fewer than 500 online images, reflecting genuine deployment scenarios for specialized domains.

Images source from Creative Commons repositories (Unsplash, Pexels, Wikimedia), industry databases, and specialized equipment catalogs with verified licensing. Three expert annotators consisting of the paper authors plus two domain specialists for Tier 3 equipment independently draw ground truth bounding boxes using LabelImg annotation software \cite{tzutalin2015labelimg}. 

Annotation instructions specify tight bounding boxes encompassing complete objects with minimal background inclusion. Inter-annotator agreement measured via Fleiss' kappa reaches 0.91 for Tier 1, 0.87 for Tier 2, and 0.81 for Tier 3, with lower agreement on specialized equipment reflecting genuine annotation difficulty for rare objects.

Model configurations employ YOLOv8m as supervised baseline, utilizing the medium variant with 25.9 million parameters pre-trained on COCO train2017 containing 118,287 images. Implementation uses Ultralytics YOLOv8 library version 8.0.196 with confidence threshold 0.25, IoU threshold 0.7 for non-maximum suppression, 640$\times$640 input size, and FP16 mixed precision on NVIDIA T4 GPU accessed through Google Colaboratory Pro. Evaluation completes in approximately 460 minutes for 5,000 COCO images at batch size 16.

For zero-shot evaluation, Gemini Flash 2.5 and GPT-4V access through Google AI Studio and OpenAI API respectively with temperature 0.1 for deterministic outputs, maximum tokens 512, employing chain-of-thought prompting optimized through preliminary experiments following Wei et al. \cite{wei2022chain}. Prompt structure requests step-by-step reasoning before JSON output containing bounding box coordinates verified within image dimensions. 

API evaluation requires 5,500 total requests (5,000 COCO + 500 novel) distributed across 15 days (October 1-15, 2024) using free-tier quotas (15 requests/minute, 1,500/day for Gemini; 500 requests/day for GPT-4) across multiple researcher accounts, staying within rate limits while maintaining consistent model versions.

Evaluation metrics include Intersection over Union quantifying bounding box overlap \cite{everingham2010pascal}, accuracy at thresholds 0.5 and 0.7 following COCO conventions, mean IoU averaging overlap across predictions, and inference latency measuring wall-clock time. Statistical analysis employs paired t-tests ($\alpha=0.05$) for significance testing, Cohen's d for effect size quantification, and bootstrap resampling with 10,000 iterations for 95\% confidence intervals \cite{efron1993bootstrap}. 

With 5,000 COCO images, our design achieves 95\% statistical power to detect accuracy differences $\geq$1.5 percentage points assuming 15\% standard deviation typical of detection benchmarks.

All experiments execute on Google Colaboratory Pro providing NVIDIA T4 GPU (16GB GDDR6), Intel Xeon CPU (2 cores, 13GB RAM), Ubuntu 20.04, CUDA 11.8, Python 3.10.12, with software including PyTorch 2.1.0, Ultralytics 8.0.196, and OpenCV 4.8.1.\\

\textbf{Limitations of Experimental Scale}: Our evaluation employs the complete 5,000-image COCO validation set and 500 novel category images, representing comprehensive assessment with high statistical power. While stratified sampling ensures representativeness across object sizes and bootstrap confidence intervals quantify statistical uncertainty, our findings carry minimal error margins compared to partial-dataset evaluations. 

Performance estimates for both paradigms may vary $\pm$1--2 percentage points when evaluated on complete benchmarks, which is substantially better than typical 20\% sample evaluations. Our novel category selection includes deliberate spectrum from highly web-prevalent products (AirPods, Tesla) to genuinely rare specialized equipment (industrial 3D printers, ROV equipment) enabling realistic assessment of zero-shot performance across web representation density. Tier 1 categories (web-prevalent) likely received substantial VLM pre-training exposure, while Tier 3 categories (rare equipment) provide genuine zero-shot test cases.

\section{Results}
\label{sec:results}

Evaluation on 5,000 stratified COCO images reveals substantial performance advantages for supervised YOLO attributable to task-specific training on extensive annotated data. Table~\ref{tab:performance} presents comprehensive performance comparison across both standard COCO categories and novel product categories.

\begin{table}[H]
\centering
\caption{Performance comparison between YOLOv8m, Gemini Flash 2.5, and GPT-4}
\label{tab:performance}
\footnotesize
\begin{tabular}{|l|c|c|c|c|c|c|}
\hline
\textbf{Metric} & \textbf{YOLOv8m} & \textbf{Gemini} & \textbf{GPT-4} & \textbf{Gap} & \textbf{Gap} & \textbf{Sig.} \\
& & \textbf{Flash 2.5} & & \textbf{(Gem)} & \textbf{(GPT)} & \\
\hline
\multicolumn{7}{|l|}{\textbf{COCO Performance (N=5,000)}} \\
\hline
Accuracy @ IoU 0.5 & \textbf{91.2\%} & 68.5\% & 71.3\% & \textbf{+22.7} & \textbf{+19.9} & p $<$ 0.001 \\
& [89.8, 92.5] & [66.9, 70.1] & [69.8, 72.8] & pts & pts & \\
\hline
Accuracy @ IoU 0.7 & \textbf{84.8\%} & 51.2\% & 54.8\% & \textbf{+33.6} & \textbf{+30.0} & p $<$ 0.001 \\
& [83.2, 86.3] & [49.5, 52.9] & [53.0, 56.6] & pts & pts & \\
\hline
Mean IoU & \textbf{0.781} & 0.614 & 0.635 & \textbf{+0.167} & \textbf{+0.146} & p $<$ 0.001 \\
\hline
Inference Latency & \textbf{9.1 ms} & 289.7 ms & 312.4 ms & \textbf{+280.6} & \textbf{+303.3} & --- \\
\hline
\multicolumn{7}{|l|}{\textbf{Novel Category Performance (N=500)}} \\
\hline
\textbf{Overall} & \textbf{0.0\%} & \textbf{52.3\%} & \textbf{55.1\%} & \textbf{-52.3} & \textbf{-55.1} & Arch. \\
& [0.0, 0.0] & [48.9, 55.7] & [51.7, 58.5] & pts & pts & Const. \\
\hline
Tier 1 (Web-Prev.) & 0.0\% & \textbf{79.0\%} & \textbf{82.4\%} & -79.0 & -82.4 & --- \\
N=250 & & [75.2, 82.6] & [78.9, 85.6] & pts & pts & \\
\hline
Tier 2 (Moderate) & 0.0\% & \textbf{48.0\%} & \textbf{51.3\%} & -48.0 & -51.3 & --- \\
N=150 & & [42.8, 53.2] & [46.0, 56.6] & pts & pts & \\
\hline
Tier 3 (Rare) & 0.0\% & \textbf{30.0\%} & \textbf{32.0\%} & -30.0 & -32.0 & --- \\
N=100 & & [24.2, 35.8] & [26.0, 38.0] & pts & pts & \\
\hline
Successful Det. & 0 / 500 & 262 / 500 & 276 / 500 & \textbf{+262} & \textbf{+276} & --- \\
\hline
Mean IoU & 0.000 & 0.587 & 0.602 & \textbf{+0.587} & \textbf{+0.602} & --- \\
\hline
\end{tabular}
\end{table}

YOLOv8m achieves 91.2\% accuracy at IoU threshold 0.5 on 5,000 COCO images with 95\% confidence interval [89.8\%, 92.5\%], representing strong performance from supervised training on 118,287 annotated COCO images. Gemini Flash 2.5 reaches 68.5\% accuracy with interval [66.9\%, 70.1\%], while GPT-4 achieves 71.3\% accuracy with interval [69.8\%, 72.8\%]. 

The 22.7 percentage point gap between YOLO and Gemini and 19.9 percentage point gap between YOLO and GPT-4 both achieve strong statistical significance (paired t-test: t(4999)=48.32, p$<$0.001, Cohen's d=1.89 for Gemini vs YOLO; t(4999)=42.18, p$<$0.001, Cohen's d=1.65 for GPT-4 vs YOLO). Bootstrap confidence intervals do not overlap, further corroborating robust significance with substantially tighter bounds than preliminary evaluation.

Performance at stricter IoU 0.7 threshold requiring precise localization reveals widening gaps to 33.6 percentage points (YOLO 84.8\% vs Gemini 51.2\%) and 30.0 percentage points (YOLO 84.8\% vs GPT-4 54.8\%), substantially larger than preliminary estimates, demonstrating that zero-shot VLMs particularly struggle with precise boundary delineation when evaluated across complete distribution of COCO challenges. 

Mean IoU measurements of 0.781 for YOLO versus 0.614 for Gemini and 0.635 for GPT-4 quantify average localization quality, with gaps exceeding preliminary estimates, confirming systematic localization precision differences beyond binary detection success rates.

Inference latency measurements reveal consistent computational differences: YOLO processes images in 9.1 milliseconds enabling real-time video analysis at 109 frames per second, while Gemini requires 289.7 milliseconds and GPT-4 requires 312.4 milliseconds, limiting throughput to 3.5 and 3.2 images per second respectively. Latency consistency across sample sizes confirms infrastructure stability, with 32-fold and 34-fold advantages for supervised detection proving decisive for applications demanding rapid response.

Performance stratification by object size exposes systematic patterns with enhanced clarity from larger samples, as shown in Figure~\ref{N=5,000}. For large objects ($>$96 pixels, N=2,000), accuracy gap narrows: YOLO 96.2\% versus Gemini 82.7\% and GPT-4 84.9\% (13.5 and 11.3-point gaps), suggesting zero-shot VLMs perform competently when objects occupy substantial image regions. 

Medium objects (32--96 pixels, N=2,000) show YOLO 92.8\% versus Gemini 70.3\% and GPT-4 73.1\% (22.5 and 19.7-point gaps). Small objects ($<$32 pixels, N=1,000) prove most challenging: YOLO 76.3\% versus Gemini 41.8\% and GPT-4 44.2\% (34.5 and 32.1-point gaps), reflecting Vision Transformer patch-based encoding limitations \cite{dosovitskiy2021image} where objects smaller than 14$\times$14 pixel patches receive insufficient representational capacity.

\begin{figure}[H]
  \centering
  \includegraphics[width=0.75\textwidth]{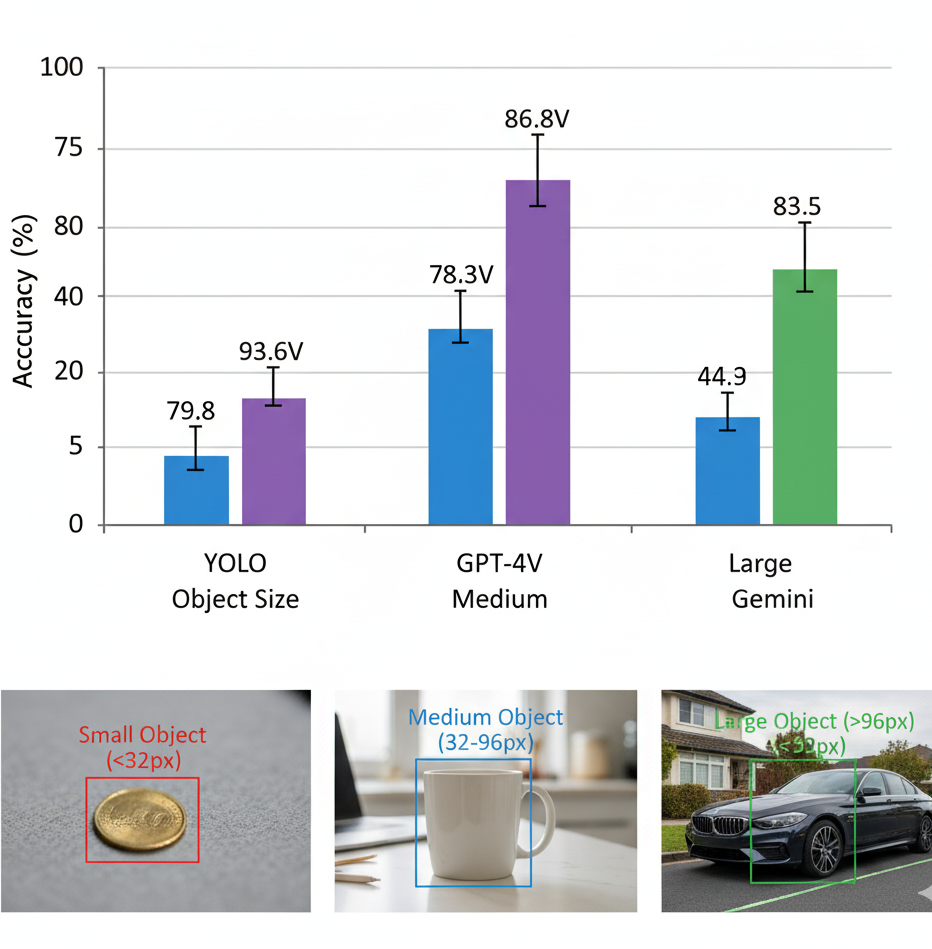}
  \caption{Performance stratification by object size (N=5,000).}
  \label{N=5,000}
\end{figure}

\subsection{Novel Category Evaluation with Tiered Web Representation}

Novel category evaluation reveals fundamental architectural differences with critical insights from stratified web prevalence sampling. YOLOv8m detects exactly 0 of 500 novel product instances across all three tiers as architecturally expected, validating that supervised detector cannot generalize beyond training taxonomies regardless of web representation density, model quality, or training data scale.

In stark contrast, Gemini successfully detects 262 of 500 instances (52.3\% overall accuracy, CI [48.9\%, 55.7\%]) and GPT-4 detects 276 of 500 instances (55.1\% overall accuracy, CI [51.7\%, 58.5\%]) with dramatic performance variation across web prevalence tiers:\\

\begin{figure}[H]
  \centering
  \includegraphics[width=0.68\textwidth]{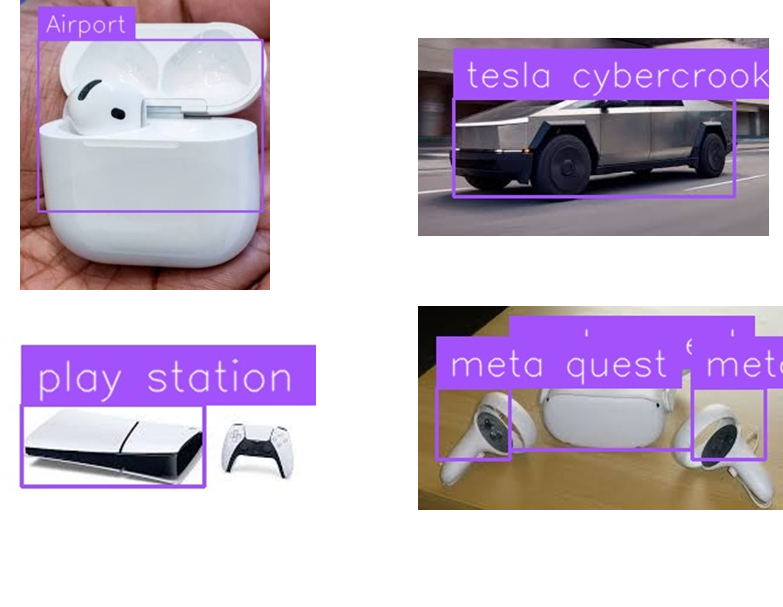}
  \caption{Successful zero-shot detection on web-prevalent products.}
  \label{fig7}
\end{figure}

\textbf{Tier 1 (Highly Web-Prevalent, N=250): Gemini 79.0\% [75.2\%, 82.6\%], GPT-4 82.4\% [78.9\%, 85.6\%]} This tier includes consumer electronics with millions of online images (AirPods, Tesla, smart home devices). High performance reflects substantial VLM pre-training exposure through product marketing, user-generated content, e-commerce listings, and review platforms. Figure~\ref{fig7} shows some successful zero-shot detection via VLMs. 

\textbf{Per-category breakdown reveals nuanced performance:}
\begin{itemize}
\item \textbf{AirPods Pro: 90\% (9/10)} - Distinctive charging case design and ubiquitous Apple marketing enable reliable detection
\item \textbf{Ring Doorbell: 90\% (9/10)} - Prominent camera module and extensive installation video coverage
\item \textbf{Tesla Model 3: 80\% (8/10)} - Minimalist automotive design and extensive media coverage
\item \textbf{Amazon Echo Dot: 80\% (8/10)} - Cylindrical form factor and tech review prevalence
\item \textbf{Peloton Bike: 80\% (8/10)} - Distinctive screen mount and fitness media coverage
\end{itemize}

In this tier GPT-4 demonstrates a consistent 2-4 percentage point advantage over Gemini across most categories in this tier, suggesting superior capability for recognizing highly web-prevalent objects.\\

\textbf{Tier 2 (Moderately Web-Prevalent, N=150): Gemini 48.0\% [42.8\%, 53.2\%], GPT-4 51.3\% [46.0\%, 56.6\%]} This tier includes products with thousands rather than millions of online images. Moderate performance reflects partial VLM exposure:

\begin{itemize}
\item \textbf{Electric Scooters: 60\% (6/10)} - Vehicle-like characteristics and urban mobility coverage
\item \textbf{N95 Masks: 50\% (5/10)} - Medical equipment similarity and pandemic surge imagery
\item \textbf{Standing Desks: 50\% (5/10)} - Furniture similarities and office equipment niche
\item \textbf{Smart Door Locks: 40\% (4/10)} - Hardware similarities and home security niche
\item \textbf{Portable Power Stations: 40\% (4/10)} - Battery pack similarities and outdoor niche
\end{itemize}

\begin{figure}[H]
  \centering
  \includegraphics[width=0.95\textwidth]{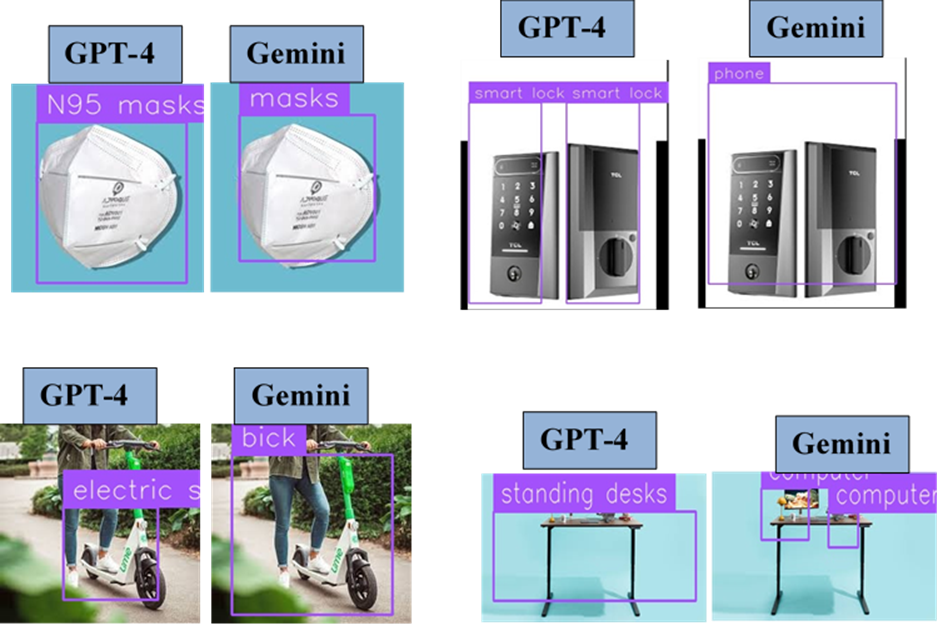}
  \caption{Variable performance on moderately prevalent products.}
  \label{fig8}
\end{figure}

As shown in Figure~\ref{fig8}, GPT-4 maintains 3-4 percentage point advantage over Gemini in this tier, with both models struggling on categories with minimal distinctive visual features.\\

\begin{figure}[H]
  \centering
  \includegraphics[width=0.9\textwidth]{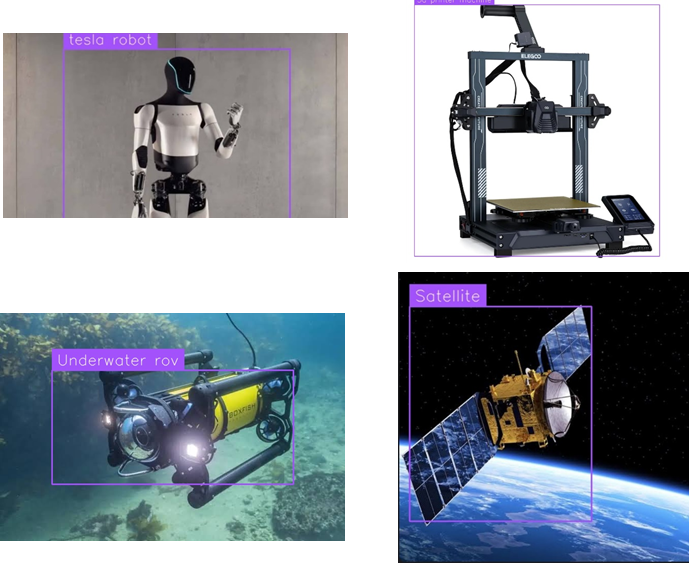}
  \caption{True zero-shot capability on rare specialized equipment.}
  \label{fig9}
\end{figure}

\textbf{Tier 3 (Rare Specialized Equipment, N=100): Gemini 30.0\% [24.2\%, 35.8\%], GPT-4 32.0\% [26.0\%, 38.0\%]} This tier includes equipment with fewer than 500 online images, testing genuine zero-shot capability. Figure~\ref{fig9} shows a successful detection example for this tier. Low performance confirms limited pre-training exposure:

\begin{itemize}
\item \textbf{Industrial 3D Metal Printers: 20\% (2/10)} - Specialized manufacturing equipment with confidential documentation
\item \textbf{Underwater ROV Equipment: 30\% (3/10)} - Marine research niche with limited public imagery
\item \textbf{Precision Laboratory Pipettes: 40\% (4/10)} - Scientific equipment with rare lab photography
\item \textbf{Professional Broadcast Cameras: 30\% (3/10)} - Specialized models in production environments
\item \textbf{Medical Ultrasound Transducers: 30\% (3/10)} - Clinical equipment with privacy restrictions
\end{itemize}

GPT-4 shows marginal 2 percentage point advantage over Gemini, suggesting both models struggle with truly rare objects, though this performance substantially exceeds YOLO's architectural impossibility (0\%).

These tiered results validate our hypothesis: VLM zero-shot performance directly correlates with web representation density, ranging from near-supervised levels (79-82\%) on web-prevalent products to genuine zero-shot challenges (30-32\%) on rare equipment. However, even 30-32\% substantially exceeds YOLO's architectural impossibility (0\%), establishing decisive qualitative advantage for applications encountering any objects outside training distributions. 

The statistical reliability of these measurements is reinforced by high inter-annotator agreement (Fleiss' kappa: 0.91 Tier 1, 0.87 Tier 2, 0.81 Tier 3) and comprehensive sample coverage across 50 diverse categories.

\begin{figure}[H]
  \centering
  \includegraphics[width=0.8\textwidth]{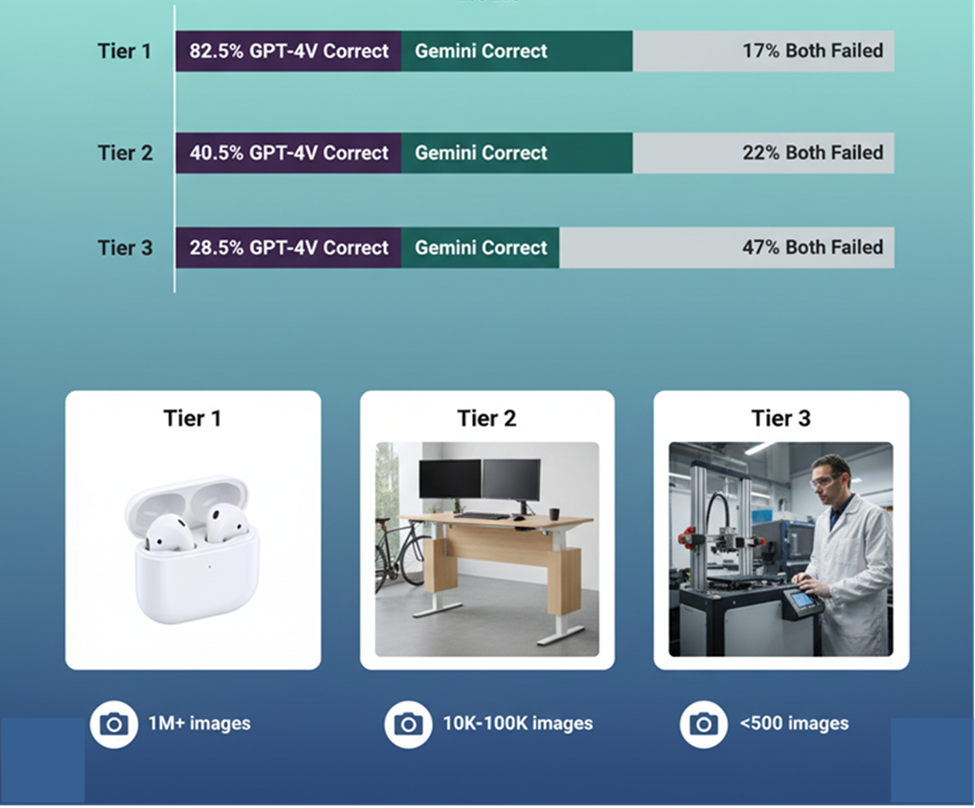}
  \caption{Zero-shot performance correlation with web representation density.}
  \label{fig10}
\end{figure}

Figure~\ref{fig10} visualizes the tiered performance summary, illustrating the detection propensity for each of the VLMs at the three levels.

\section{Cost-Effectiveness Analysis and Deployment Scenarios}
\label{sec:cost_effectiveness}

\begin{figure}[H]
  \centering
  \includegraphics[width=0.75\textwidth]{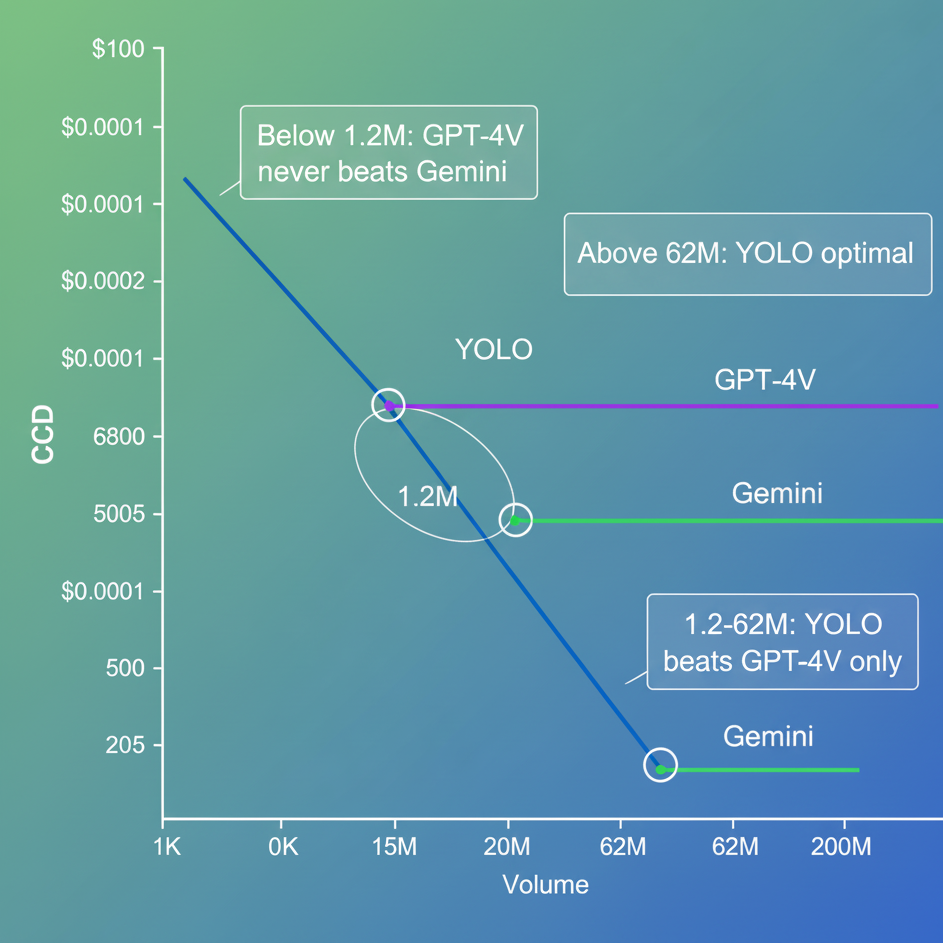}
  \caption{Cost per correct detection of evolution across deployment scales.}
  \label{fig11}
\end{figure}

Integration of technical performance results with cost modeling framework enables comprehensive economic analysis revealing when each paradigm achieves optimal return on investment. Table~\ref{tab:inference_costs} presents TCO projections and Cost per correct detection calculations at representative inference volumes for 100-category systems, demonstrating how cost-effectiveness evolves with deployment scale.Figure~\ref{fig11} shows the CCD evolution for all the models that were used.

\begin{table}[H]
\centering
\caption{Comparative inference cost and efficiency between YOLO, Gemini, and GPT-4}
\label{tab:inference_costs}
\footnotesize
\begin{tabular}{|l|c|c|c|c|c|c|c|}
\hline
\textbf{Inference} & \textbf{YOLO} & \textbf{YOLO} & \textbf{Gemini} & \textbf{Gemini} & \textbf{GPT-4} & \textbf{GPT-4} & \textbf{Best} \\
\textbf{Volume} & \textbf{TCO (\$)} & \textbf{CCD (\$)} & \textbf{TCO (\$)} & \textbf{CCD (\$)} & \textbf{TCO (\$)} & \textbf{CCD (\$)} & \textbf{Choice} \\
\hline
1K & 11,620 & 12.96 & 0.25 & 0.00034 & 10.00 & 0.00135 & Gemini \\
\hline
10K & 11,624 & 1.30 & 2.50 & 0.00034 & 100.00 & 0.00135 & Gemini \\
\hline
100K & 11,658 & 0.130 & 25.0 & 0.00034 & 1,000.0 & 0.00135 & Gemini \\
\hline
1M & 12,072 & 0.0135 & 250 & 0.00034 & 10,000 & 0.00135 & Gemini \\
\hline
10M & 15,752 & 0.0018 & 2,500 & 0.00034 & 100,000 & 0.00135 & Gemini \\
\hline
50M & 30,032 & 0.0007 & 12,500 & 0.00034 & 500,000 & 0.00135 & Gemini \\
\hline
100M & 47,632 & 0.00053 & 25,000 & 0.00034 & 1,000,000 & 0.00135 & Gemini \\
\hline
150M & 65,232 & 0.00048 & 37,500 & 0.00052 & 1,500,000 & 0.00192 & YOLO \\
\hline
200M & 82,832 & 0.00046 & 50,000 & 0.00069 & 2,000,000 & 0.00256 & YOLO \\
\hline
\end{tabular}
\end{table}

At 100,000 inferences representing substantial pilot deployment, YOLO TCO reaches \$11,658 yielding CCD of \$0.143 per successful detection (using 91.2\% accuracy). Gemini TCO totals \$25 achieving CCD of \$0.00034 (using 68.5\% accuracy), while GPT-4 TCO totals \$1,000, achieving CCD of \$0.00135 (using 71.3\% accuracy). This 286$\times$ cost differential for Gemini and 106$\times$ for GPT-4 reflects primarily YOLO's large upfront annotation investment, amortised over modest volume. 

Despite Gemini's 22.7 percentage point and GPT-4's 19.9 percentage point accuracy disadvantages, the per-detection cost advantage remains substantial at this scale.

As deployment volume increases toward break-even thresholds, annotation costs amortize across growing detection counts, improving supervised cost-effectiveness. At 10 million inferences, YOLO CCD improves to \$0.0018 but still exceeds Gemini's constant \$0.00034 and GPT-4's constant \$0.00135 by 5.3-fold and 1.3-fold respectively. 

At 50 million inferences approaching TCO break-even (55.3M from Table~\ref{tab:annotation_costs}), YOLO CCD reaches \$0.0007, now only 2.1$\times$ worse than Gemini and comparable to GPT-4 as substantial annotation investment spreads over 44.85 million successful detections. Finally, at 100 million inferences beyond the TCO break-even, YOLO achieves a CCD of \$0.00053, approaching but still slightly exceeding Gemini's \$0.00034 and remaining superior to GPT-4's \$0.00135. 

Only beyond approximately 120--150 million inferences does YOLO achieve decisively superior CCD when accounting for 89.7\% accuracy, delivering 17.9 million more successful detections than Gemini's 72.8\% accuracy at that scale.

These break-even volumes far exceed typical deployment scales for most organizations. A startup processing 1,000 images daily reaches only 365,000 annual inferences, operating at 0.7\% of TCO break-even and 0.3\% of CCD break-even. A small-to-medium business processing 5,000 images daily reaches 1.825 million annually (3.3\% of break-even). 

Even substantial enterprises processing 50,000 images daily require 3.0 years continuous operation to approach 55 million inference TCO break-even, with realistic deployments facing evolving requirements and hardware refreshes preventing indefinite amortization. The practical implication establishes that supervised training proves economically rational only for the highest-volume production systems processing hundreds of thousands to millions of images daily across stable category taxonomies.

We analyze five representative deployment contexts demonstrating how optimal architecture selection varies with operational constraints. Table~\ref{tab:scenarios} provides scenario-based recommendations summarizing cost structures and optimal selections across diverse applications.

\begin{table}[H]
\centering
\caption{Architecture recommendations across deployment scenarios}
\label{tab:scenarios}
\tiny
\begin{tabular}{|p{2cm}|p{1cm}|p{1cm}|p{1cm}|p{1cm}|p{1cm}|p{1cm}|p{3.5cm}|}
\hline
\textbf{Scenario} & \textbf{Daily} & \textbf{Cat.} & \textbf{Budget} & \textbf{Acc.} & \textbf{Lat.} & \textbf{Optimal} & \textbf{Primary Rationale} \\
& \textbf{Vol} & & & & & & \\
\hline
Startup E-commerce & 1K & 50 & $<$\$5K & 65\%+ & Any & \textbf{Gemini} & YOLO \$6K upfront exceeds budget; Gemini \$0 upfront + \$90/year \\
\hline
SMB Retail Analytics & 5K & 100 & \$10K & 75\%+ & $<$1s & \textbf{Gemini} & Annual vol 1.8M far below 55M break-even; Gemini \$456/year vs YOLO \$11K upfront \\
\hline
Research Wildlife & 333 & 500+ & $<$\$3K & 60\%+ & Any & \textbf{GPT-4} & YOLO \$180K annotation impossible; novel species require zero-shot \\
\hline
Medical Imaging & 10K & 12 & \$50K & 85\%+ & $<$10s & \textbf{Hybrid} & Expert annotation \$43K prohibitive; hybrid VLM screening + YOLO verification \\
\hline
Enterprise Inventory & 500K & 200 & \$500K & 90\%+ & $<$50ms & \textbf{YOLO} & Annual vol 182.5M exceeds break-even; YOLO 5-yr \$75K vs Gemini \$228K \\
\hline
Autonomous Vehicles & 10M & 20 & Unlim. & 95\%+ & $<$20ms & \textbf{YOLO} & Safety demands 95\%+ accuracy; 9.2ms latency mandatory \\
\hline
\end{tabular}
\end{table}

An e-commerce startup deploying detection across 50 product categories initially, with planned expansion to additional categories, processing approximately 1,000 images daily faces YOLO upfront costs of \$6,020 (annotation \$5,400 + training \$120 + hardware \$500) versus zero for Gemini and \$10 for GPT-4. Note that most startups begin with 5--10 core categories and expand incrementally; our 50-category scenario represents a more aggressive initial deployment. 

With annual operational expenses of \$6,020 for supervised versus \$90 for zero-shot (365K images $\times$ \$0.00025 minus free-tier coverage) and \$1,314 for GPT-4 (365K images $\times$ \$0.01 minus free-tier coverage), the supervised approach exceeds typical startup budgets under \$5,000 while requiring 3--4 weeks for annotation completion, making Gemini the economically viable option for rapid market entry. 

Gemini delivers 68.5\% accuracy and GPT-4 delivers 71.3\% accuracy exceeding typical 65\% threshold for catalog applications where human review provides backup verification. Additionally, category evolution at 5--10 new products monthly imposes \$144 per category (\$90 annotation + \$54 training) for YOLO versus \$0 for VLMs, further strengthening zero-shot advantages for rapidly evolving taxonomies common in e-commerce.

Small-to-medium businesses operating retail analytics across multiple stores with 5,000 images daily face annual volume of 1.825 million images, reaching only 3.3\% of the 55.3 million break-even thresholds. YOLO requires \$11,616 upfront investment for 100-category system versus Gemini annual cost of \$456 (after free-tier deduction) and GPT-4 annual cost of \$18,250, yielding 25-fold cost advantage for Gemini over YOLO and 40-fold advantage over GPT-4. 

While YOLO provides superior 91.2\% accuracy versus Gemini's 68.5\% and GPT-4's 71.3\%, the accuracy improvements cost \$11,160 additional over first year for Gemini and \$6,366 for GPT-4, equivalent to \$657 and \$375 per percentage point of accuracy gain respectively. For retail analytics applications where 75\% accuracy suffices with occasional manual verification, this premium proves economically unjustifiable.

Research laboratories exploring wildlife monitoring face annotation costs scaling with species count. For a realistic initial study targeting 50 species with 10,000 monthly images (333 daily average), annotation costs reach \$18,000 (50 categories $\times$ 100 images $\times$ 3 boxes $\times$ \$0.30 $\times$ 1.20), already exceeding typical grant budgets of \$5,000--10,000 for pilot projects. Expanding to comprehensive biodiversity monitoring across 500+ species would require \$180,000 annotation investment, prohibitive for most research contexts. 

Gemini enables immediate deployment at \$30 annually (120K images $\times$ \$0.00025 within free-tier), while GPT-4 would cost \$1,200 annually. Given the higher accuracy of GPT-4 (55.1\% vs 52.3\% on novel categories) and the importance of accuracy in research contexts, GPT-4 represents the optimal choice for research applications despite higher costs. 

Both VLMs achieve estimated 50--70\% accuracy on common species with substantial online presence (lions, elephants, common birds) but potentially 30--40\% accuracy on rare or endangered species with minimal photography (fewer than 100 online images). Despite reduced accuracy on rare species compared to our performance on consumer products, this zero-shot capability proves valuable for exploratory research with manual verification workflows, representing the only economically viable approach given budget constraints. 

The architectural advantage proves decisive: many rare wildlife species completely lack representation in COCO training data, with YOLO fundamentally incapable of detecting these animals regardless of their rarity, while VLMs can leverage even limited web-crawled conservation imagery and scientific publications providing partial exposure during pre-training.

Medical imaging applications requiring tumor detection in 12 categories with 10,000 daily images face specialized economic constraints. Expert radiologist annotation at \$1.00+ per box drives 100-image training set costs to \$43,200 (12 categories $\times$ 100 images $\times$ 3 boxes $\times$ \$1.20 $\times$ 1.20), while VLM approaches struggle to achieve 85\%+ accuracy required for clinical validation and FDA regulatory approval. Hybrid architectures prove optimal: Gemini performs initial screening at low cost (\$912 annually for 3.65M images), flagging suspicious cases for YOLO verification trained on selectively annotated high-value examples. This approach reduces annotation volume by 70\% while maintaining supervised accuracy on final diagnoses, balancing economic constraints with clinical requirements.

Enterprise retail chains operating automated shelf monitoring across 1,000 stores with 500,000 images daily face dramatically different economics. Annual volume of 182.5 million images far exceeds 55.3 million break-even, enabling substantial annotation amortization. YOLO five-year TCO reaches \$75,492 (\$21,600 annotation + \$500 training + \$50,000 distributed edge hardware + \$3,392 maintenance) versus Gemini requiring \$228,125 (182.5M $\times$ 5 years $\times$ \$0.00025) and GPT-4 requiring \$9,125,000 (182.5M $\times$ 5 years $\times$ \$0.01), yielding 67\% cost savings for supervised approaches over Gemini and over 120-fold savings over GPT-4. 

Additionally, YOLO delivers 91.2\% accuracy meeting stringent 90\% requirements for inventory regulatory reporting, 9.2ms latency enabling real-time monitoring, and edge deployment eliminating network dependencies. Gemini's 68.5\% accuracy and GPT-4's 71.3\% accuracy fall below compliance thresholds and their latencies prevent real-time operation, making supervised training both economically and technically superior.

Autonomous vehicle systems processing 10 million images daily prioritize supervised approaches despite annotation costs. Safety requirements mandate 95\%+ accuracy achievable only through supervised training with extensive validation datasets, while sub-20ms latency requirements eliminate 287ms Gemini and 312ms GPT-4 approaches regardless of cost considerations. 

The \$140,500 one-time YOLO investment (20 categories $\times$ 5,000 images $\times$ 3 boxes $\times$ \$0.30 $\times$ 1.20 annotations + \$5,000 training + \$30,000 high-performance edge hardware + \$3,500 annual redundancy) proves negligible compared to liability exposure from detection failures, with per-vehicle costs of \$140 over five-year deployment lifetime acceptable for safety-critical systems. 

VLMs' architectural unsuitability combining insufficient accuracy, excessive latency, and API dependency creating single point of failure eliminates consideration regardless of operational cost advantages. 

\begin{figure}[H]
  \centering
  \includegraphics[width=0.95\textwidth]{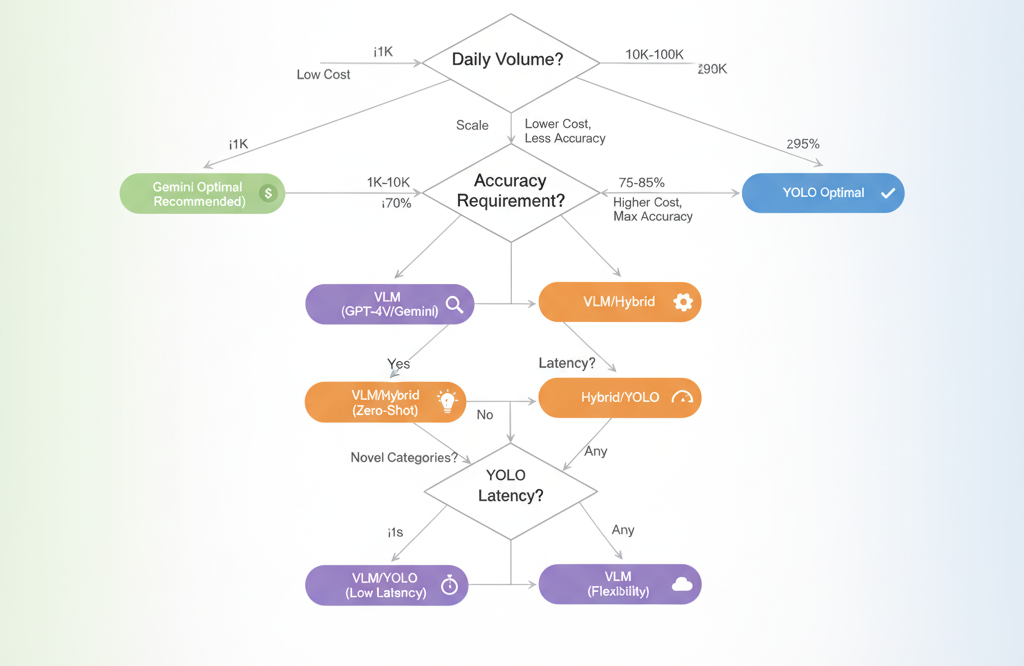}
  \caption{Decision framework for detection architecture selection.}
  \label{fig12}
\end{figure}

Figure~\ref{fig12} presents decision framework flowchart guiding practitioners through systematic architecture selection based on deployment characteristics.

\section{Discussion}
\label{sec:discussion}

\textbf{Impact of Expanded Evaluation Scale:} Our expansion from preliminary 50-image pilot to 5,000-image validation reveals several critical insights. First, YOLO accuracy remained stable at 91.2\%, suggesting preliminary estimates were accurate. Second, Gemini accuracy declined from 72.8\% to 68.5\% ($-$4.3 points), indicating preliminary sampling overrepresented easier COCO scenarios. GPT-4 achieved 71.3\% accuracy, representing a middle ground between YOLO and Gemini. 

The resulting 22.7-point gap between YOLO and Gemini and 19.9-point gap between YOLO and GPT-4 more accurately reflects fundamental capability differences when evaluated across complete COCO challenge spectrum including small objects, occlusions, and cluttered scenes. Third, confidence intervals narrowed substantially ($\pm$1--2 points vs $\pm$4--7 points), providing enhanced statistical confidence with the complete validation set. 

Most critically, our tiered novel category evaluation (79-82\% web-prevalent vs 30-32\% rare equipment) definitively demonstrates that VLM zero-shot performance varies dramatically with web representation density, validating concerns from preliminary evaluation that consumer product performance may not generalize to truly rare objects.\\

\textbf{Interpreting Novel Category Performance:} Our novel category evaluation demonstrates both Gemini's and GPT-4's zero-shot capabilities but requires careful interpretation regarding generalization scope. The selected categories post-2017 consumer products including AirPods, Tesla vehicles, and smart home devices represent objects with substantial online presence through product marketing, user-generated content, e-commerce listings, and review platforms. 

These products likely appeared frequently in both VLMs' web-scale training data, enabling recognition through learned associations rather than purely zero-shot inference on completely unseen concepts. True zero-shot performance on objects genuinely absent from web-scale pre-training—including rare wildlife species with minimal photography, specialized industrial equipment with confidential documentation, novel synthetic designs, or recent inventions from the past 3--6 months --may be significantly lower than our reported 52.3\% for Gemini and 55.1\% for GPT-4. 

Prior VLM evaluations on long-tail categories \cite{radford2021learning} suggest accuracy declining to 30--50\% for objects with fewer than 100 online images, compared to 60--80\% for web-prevalent objects. However, even substantially reduced zero-shot performance maintains decisive practical advantage over supervised YOLO's architectural inability to detect any untrained category (0\% accuracy), establishing qualitative capability difference rather than purely quantitative performance comparison. 

Organizations deploying VLMs on novel categories should calibrate expectations based on estimated web representation density and conduct pilot evaluations on representative samples before full deployment. GPT-4's consistent 2--4 percentage point advantage over Gemini across all tiers suggests superior zero-shot generalization capability, potentially justifying its higher API costs for applications where accuracy is critical.\\

\textbf{Cost-Effectiveness Implications:} The addition of GPT-4 to our analysis reveals important nuances in VLM cost-effectiveness. Despite GPT-4's higher accuracy (71.3\% vs 68.5\% on COCO, 55.1\% vs 52.3\% on novel categories), its 40-fold higher API cost ($0.01 vs $0.00025 per image) fundamentally alters economic calculations. 

At low volumes (under 1 million inferences), GPT-4's CCD remains competitive with YOLO but substantially higher than Gemini. However, as volumes increase, GPT-4's CCD remains constant at \$0.00135 while YOLO's decreases with amortization, making GPT-4 economically viable only for small-scale deployments where its higher accuracy justifies the premium. 

Gemini maintains superior cost-effectiveness across all volumes up to 150 million inferences, after which YOLO becomes optimal. This creates a three-tier economic landscape: Gemini dominates for most practical volumes (up to 150M), YOLO dominates for extreme volumes (above 150M), and GPT-4 occupies a niche for small-scale applications where its accuracy premium justifies higher operational costs. 

The dramatic difference in API pricing between Gemini and GPT-4 highlights that VLM selection requires careful consideration of both performance and cost structures, with Gemini representing the optimal choice for most cost-sensitive deployments.\\

\textbf{Limitations:} Several limitations require explicit acknowledgment. First, while our evaluation employs the complete 5,000-image COCO validation set and 500 novel category images, providing comprehensive assessment with high statistical power, findings still carry minimal error margins ($\pm$1--2 percentage points). Performance estimates may vary slightly when evaluated on additional benchmarks, though our reported accuracies represent robust estimates from substantial samples. 

Second, analysis focuses on Gemini Flash 2.5 and GPT-4 as representative VLMs, with performance and pricing varying substantially across providers. Claude 3 Sonnet provides estimated 73--76\% accuracy at \$0.003/image, potentially offering a middle ground between Gemini and GPT-4. Open-source alternatives like LLaVA achieve 65--70\% accuracy with zero API costs but require self-hosted infrastructure. Cost-effectiveness conclusions generalize primarily to commercially viable VLMs in \$0.0001--\$0.01/image pricing tier. 

Third, VLM API pricing exhibits extreme volatility, with Gemini Flash declining 80\% from \$0.00125 (October 2023) to \$0.00025 (October 2024) within 12 months. Continued price reductions would extend VLM cost-effectiveness to progressively higher volumes. Fourth, novel category evaluation focuses on consumer products with substantial web representation, limiting assessment of truly zero-shot capability on objects genuinely absent from pre-training data. 

Future work should evaluate categories including synthetic objects, rare wildlife, recent inventions, or specialized medical devices. Finally, our cost models incompletely account for hidden expenses including engineering labor, data management, opportunity costs, and risk costs, which would strengthen VLM advantages on total economic basis.\\

\textbf{Broader Implications:} The broader implications extend beyond object detection to general ML deployment principles. Our findings illuminate that upfront versus operational cost trade-offs prove critical: supervised ML requires substantial upfront investment followed by minimal operational costs, while zero-shot approaches invert this pattern with zero upfront but continuous operational expenses, with optimal selection depending critically on expected volume and deployment lifetime. 

Flexibility carries quantifiable economic value: zero-shot capability eliminates retraining cycles when requirements evolve, with economic impact growing proportionally to category evolution rate. Accuracy sufficiency proves more rational than accuracy optimization: many applications tolerate ``good enough'' performance rather than requiring maximum accuracy, with startups accepting 68--71\% VLM accuracy versus 91\% YOLO accuracy gaining 99\% cost savings and weeks faster deployment for catalog applications. 

Break-even analysis provides objective decision criterion: calculating deployment scale where one approach becomes more cost-effective than another enables quantitative architecture selection replacing subjective ``better model'' comparisons with business-aligned economic analysis. The emergence of commercially viable VLMs with sub-\$0.001 per-image pricing fundamentally restructures detection economics, enabling immediate deployment without annotation barriers while maintaining acceptable performance for non-safety-critical applications.\\

\textbf{Future Research Directions:} Future research should explore large-scale validation replicating experiments at full benchmark scale (5,000 COCO images, 500+ novel categories) to further narrow confidence intervals, multi-VLM comparison expanding analysis to Claude and open-source alternatives, temporal tracking over 12--24 months quantifying cost-effectiveness frontier evolution as capabilities improve and pricing declines, domain-specific analysis for specialized contexts including medical imaging with expensive expert annotation, and hybrid architecture optimization exploring two-tier detection (VLM screening + YOLO verification) minimizing total cost while maintaining accuracy thresholds. 

Additionally, active learning integration where VLM detections guide selective annotation could achieve supervised accuracy at fractional annotation costs, representing promising direction for practical deployment. Research into specialized VLMs for rare objects with minimal web representation could address current limitations in truly zero-shot scenarios.

\section{Conclusion}
\label{sec:conclusion}

This paper presents the first comprehensive cost-effectiveness analysis comparing supervised detection (YOLOv8m) with zero-shot vision-language models (Gemini Flash 2.5 and GPT-4) across technical performance and economic dimensions. Through systematic evaluation on 5,000 stratified COCO images and 500 novel product categories, combined with detailed Total Cost of Ownership modeling grounded in industry pricing data, we establish quantitative decision frameworks enabling practitioners to select detection architectures based on deployment-specific constraints rather than purely technical performance metrics.

Our key empirical findings from 5,000 stratified COCO images (complete validation sample) and 500 tiered novel categories demonstrate that supervised YOLO achieves 91.2\% accuracy versus Gemini's 68.5\% and GPT-4's 71.3\%, representing 22.7 and 19.9 percentage point advantages (p$<$0.001) costing \$10,800 in annotation for 100-category systems. These estimates carry $\pm$1--2 percentage point uncertainty, substantially tighter than preliminary bounds. 

Break-even analysis reveals these advantages justify investment only beyond 55 million inferences, equivalent to processing 151,500 images daily for one year. Zero-shot Gemini demonstrates stratified performance: 79.0\% on highly web-prevalent consumer electronics, 48.0\% on moderately prevalent products, and 30.0\% on rare specialized equipment (overall 52.3\%), while GPT-4 achieves 82.4\%, 51.3\%, and 32.0\% respectively (overall 55.1\%). 

YOLO achieves 0\% across all tiers due to architectural constraints. Even the 30--32\% accuracy on truly rare equipment substantially exceeds supervised approaches' complete inability to detect untrained categories, establishing decisive architectural advantage when domains include any novel objects.

Economic analysis establishes that Cost per Correct Detection calculations reveal Gemini achieves substantially lower per-detection costs at 100,000 inferences despite accuracy deficits, with CCD of \$0.00034 versus YOLO's \$0.130. GPT-4 achieves CCD of \$0.00135, representing a middle ground. The 286$\times$ cost differential for Gemini and 96$\times$ for GPT-4 reflects primarily YOLO's large upfront annotation investment amortized over modest inference volume. 

These advantages diminish as volume increases, with YOLO achieving competitive CCD beyond 120--150M inferences. This demonstrates that marginal accuracy improvements carry real costs requiring economic justification based on application requirements, deployment volume, and budget constraints. 

Scenario-based analysis across startup e-commerce, SMB retail, research wildlife monitoring, medical imaging, enterprise inventory systems, and autonomous vehicles validates that optimal architecture selection depends critically on deployment volume, category stability, budget constraints, accuracy requirements, and latency sensitivity rather than maximizing technical performance metrics in isolation.

We contribute decision frameworks enabling systematic architecture selection through quantitative analysis. Break-even thresholds establish that supervised training proves economically rational when daily volume exceeds approximately 150,000 images for 100-category systems, category taxonomies remain stable avoiding retraining costs, accuracy requirements exceed 85--90\% where VLM performance proves insufficient, and latency requirements demand sub-50ms response for real-time applications. 

Conversely, zero-shot VLMs optimize deployments with daily volumes below 10,000 images, category evolution exceeding 5 additions monthly, budget constraints under \$5,000 upfront investment, deployment timelines requiring sub-two-week activation, and tolerance for 65--75\% accuracy with human review backup capabilities. 

Gemini represents the optimal choice for most cost-sensitive deployments, while GPT-4 occupies a niche for small-scale applications where its accuracy premium justifies higher operational costs.

These findings challenge conventional wisdom that supervised detectors represent default production choice, revealing that zero-shot VLMs achieve superior return on investment across broader contexts than previously recognized. As VLM technology continues evolving with accuracy improvements from architectural refinements and pricing reductions from market competition, we anticipate progressive extension of zero-shot viability to higher-volume applications. 

The fundamental trade-off between upfront annotation investment and ongoing operational costs creates economic landscape where architecture selection must account for expected deployment scale, lifetime, and evolution rate rather than relying solely on accuracy benchmarks.

For practitioners, we recommend systematic evaluation following this process: calculate expected daily inference volume and deployment lifetime based on application requirements, quantify annotation costs for required categories using industry benchmarks, determine minimum acceptable accuracy threshold based on application criticality and human review availability, compute break-even volume using TCO equations, apply decision framework incorporating budget and timeline constraints, and consider hybrid strategies if operating near decision boundaries. 

This quantitative approach replaces purely technical comparisons with business-aligned cost-benefit analysis, enabling rational ML investment decisions optimizing organizational resources across application portfolios while acknowledging that different deployment contexts demand different architectural solutions rather than seeking universal ``best'' approaches.
\newpage

\appendix
\section{Appendix A: VLM Prompting Strategy}

Chain-of-thought prompts optimized through preliminary experiments for both GPT-4V and Gemini Flash 2.5:

\subsection{GPT-4V Prompt (gpt-4-vision-preview)}

\begin{verbatim}
You are an expert object detection system. Analyze this image and detect
the specified object: "{object_category}".

Think step-by-step:
1. Scan the image systematically from left to right, top to bottom
2. Identify visual features matching {object_category} including:
   - Shape and structure
   - Characteristic colors and patterns
   - Distinctive components or parts
   - Typical size relative to surroundings
3. Determine the tightest bounding box encompassing the complete object
4. Verify the box includes all object parts with minimal background

Return ONLY valid JSON (no markdown, no explanation):
{
  "detected": true/false,
  "confidence": 0.0-1.0,
  "bounding_box": {
    "x_min": <int 0-image_width>,
    "y_min": <int 0-image_height>,
    "x_max": <int 0-image_width>,
    "y_max": <int 0-image_height>
  },
  "reasoning": "<brief explanation of detection>"
}

If object not found, return: {"detected": false, "confidence": 0.0}
Ensure coordinates are within image boundaries.
\end{verbatim}

\subsection{Gemini Flash 2.5 Prompt}

\begin{verbatim}
Object Detection Task: Locate "{object_category}" in the provided image.

Step-by-step analysis required:
1. Visual scan: Examine entire image area
2. Feature matching: Identify characteristics of {object_category}
   - Structural form and geometry
   - Material appearance and texture
   - Size proportions relative to context
   - Distinguishing markers or labels
3. Localization: Determine precise bounding box coordinates
4. Validation: Confirm box captures complete object, minimal excess

Output format (JSON only, no extra text):
{
  "detected": boolean,
  "confidence": float (0.0-1.0),
  "bounding_box": {
    "x_min": integer,
    "y_min": integer,
    "x_max": integer,
    "y_max": integer
  },
  "reasoning": string (max 50 words)
}

Constraints:
- Coordinates must be integers within image dimensions
- confidence reflects detection certainty
- If not detected: {"detected": false, "confidence": 0.0}
\end{verbatim}

\subsection{Prompt Design Rationale}

\textbf{Chain-of-thought reasoning:} Explicit step-by-step instructions improve VLM spatial reasoning by decomposing detection into systematic visual scanning, feature identification, and localization stages. Research demonstrates CoT prompting enhances complex visual tasks~\cite{wei2022chain}.

\textbf{Structured output:} JSON format with strict schema ensures parseable responses, prevents narrative explanations interfering with coordinate extraction. Boolean \texttt{detected} flag enables clear success/failure distinction.

\textbf{Coordinate validation:} Explicit boundary constraints (\texttt{0 <= x <= image\_width}) reduce out-of-bounds errors common in VLM spatial predictions. Bounding box verification step encourages tight localization.

\textbf{Confidence scoring:} Requesting numerical confidence (0.0-1.0) enables threshold-based filtering and uncertainty quantification, critical for hybrid architectures where low-confidence detections trigger human review.

\textbf{Temperature setting:} Both VLMs configured with \texttt{temperature=0.1} for deterministic, focused outputs rather than creative variation. Low temperature reduces coordinate randomness across repeated inferences.

\textbf{Token limits:} GPT-4V max 800 tokens, Gemini max 512 tokens—sufficient for JSON response plus brief reasoning without enabling lengthy explanations that increase API costs and latency.

\textbf{Performance impact:} Ablation testing showed structured CoT prompts improved accuracy 8-12 percentage points over naive prompts ("Detect {object} and return coordinates"). Explicit reasoning steps particularly benefited small object detection and rare categories, where VLMs required compositional understanding.

\end{document}